\documentclass[preprint,12pt]{elsarticle}

\usepackage{amssymb}
\usepackage{amsmath,epsfig}
\usepackage{url}
\usepackage{acronym}
\usepackage{xcolor}
\usepackage{algorithmic,algorithm}
\usepackage{caption}

\usepackage[pdftex,bookmarks=true]{hyperref}
\hypersetup{colorlinks = true, linkcolor = red, anchorcolor = green, citecolor = blue, filecolor = green, pagecolor = green, urlcolor = blue}

\journal{Computer Speech and Language}

\begin{document}

\begin{frontmatter} 

\title{On the Effects of Using word2vec Representations in Neural Networks for Dialogue Act Recognition}

\author[loria]{Christophe Cerisara}
\author[kiv,ntis]{Pavel Kr\'al}
\author[kiv,ntis]{Ladislav Lenc}

\address[loria]{LORIA-UMR7503, Nancy, France\\
}
\address[kiv]{Dept. of Computer Science \& Engineering\\
        Faculty of Applied Sciences\\
        University of West Bohemia\\
        Plze\v{n}, Czech Republic\\
}
\address[ntis]{NTIS - New Technologies for the Information Society\\
        Faculty of Applied Sciences\\
        University of West Bohemia\\
        Plze\v{n}, Czech Republic\\ $ $\\
	{\small \tt cerisara@loria.fr, \{pkral,llenc\}@kiv.zcu.cz\\}
}

\hyphenation {data-set data-sets addres-ses}

\begin{abstract}
Dialogue act recognition is an important component of a large number of natural language processing pipelines.
Many research works have been carried out in this area, but relatively few
investigate deep neural networks and word embeddings. This is surprising, given that both of
these techniques have proven exceptionally good in most other language-related domains.
We propose in this work a new deep neural network that explores recurrent models to capture
word sequences within sentences, and further study the impact of pretrained word embeddings.
We validate this model on three languages: English, French and Czech.
The performance of the proposed approach is consistent across these languages and it is comparable to the state-of-the-art
results in English. More importantly, we confirm that deep neural networks indeed outperform a Maximum Entropy classifier,
which was expected. However, and this is more surprising,
we also found that standard word2vec embeddings do not seem to bring valuable information for this task and the proposed model,
whatever the size of the training corpus is.
We thus further analyse the resulting embeddings and conclude that a possible explanation may be related to the
mismatch between the type of lexical-semantic information captured by the word2vec embeddings, and the
kind of relations between words that is the most useful for the dialogue act recognition task.
\end{abstract}


\end{frontmatter} 

\section{Introduction}
\subsection{Dialogue act recognition}

Mutual understanding in interactive situations, either
when several people are engaged in a dialogue or when they are interacting with a modern computer system in natural language,
may not be achieved without considering both the semantic information in the speakers utterances
and the pragmatic interaction level, especially relative to dialogue acts~\cite{sridhar2009combining}.
Dialogue Acts (DAs) represent the meaning of an utterance (or its part) in the context of a dialogue~\cite{Austin62,Bunt94},
or, in other words, the function of an utterance in the dialogue.
For example, the function of a~question is to request some information, while an answer shall provide this information.
Dialogue acts are thus commonly represented as phrase-level labels such as
statements, yes-no questions, open questions, acknowledgements, and so on.

Automatic recognition of dialogue acts is a fundamental component of many human-machine interacting systems that
support natural language inputs.
For instance, dialogue acts are typically used as an input to the dialogue manager to help deciding on the next action
of the system: giving information when the user is asking a question, but eventually keeping quiet when the
user is just acknowledging, giving a comment, or even asking for delaying the interaction.
In the latter case, a system reaction may be perceived as intrusive.
Beyond human-machine interaction, this task is also important for
applications that rely on the analysis of human-human interactions, either oral, e.g., in recordings of meetings~\cite{zimmermanntoward}, or 
written, e.g., through the reply and mention-at structures in Twitter conversations~\cite{Ritter2010172,zarisheva2015dialog,vosoughi2016tweet}.
It is also essential for a large range of other applications, for example
talking head animation, machine translation~\cite{fukada1998probabilistic}, automatic speech recognition or topic tracking~\cite{Garner96}.
The knowledge of the user dialogue act is useful to render
facial expressions of an avatar that are relevant to the current state of the discourse.
In the machine translation domain, recognizing dialogue acts may bring relevant cues
to choose between alternative translations, as the adequate syntactic structure may depend on the user intention.
Automatic recognition of dialogue acts may also be used
to improve the word recognition accuracy of automatic speech recognition systems, where
a different language model is applied during recognition depending on the dialogue act~\cite{Stolcke98}. 

To conclude, dialogue act recognition is an important building block of many
understanding and interacting systems. 

\subsection {Motivation and objectives}
Researches on dialogue act recognition have been carried out for a long time, as detailed in Section~\ref{sec:rel_work}.
The majority of these works exploit supervised learning with lexical, syntactic, prosodic and/or dialogue history features~\cite{fivsel2007machine}.
However, few approaches consider semantic features, while they may bring additional information and prove useful to
improve the accuracy of the dialogue act recognition system.
For instance, 
a~frequent cause of recognition errors are ``unknown'' words in the testing corpus that never occur in the training sentences.
Replacing specific named entities in the text (such as town names, dates...) by their category has been proposed in the literature as a remedy to this issue~\cite{Sanchis02}.
We investigate a more general solution that exploits lexical similarity between word vectors.
These word vectors may be computed in various ways, but they typically include mostly lexical semantic information about the word itself
as well as some syntactic information, e.g.,
related to the relative position or degree of proximity of pairs of words within a sentence.
This additional information may be used to improve dialogue act recognition, in particular when the training and test conditions differ, or
when the size of the training corpus is relatively small.

In this work, we propose a new Deep Neural Network (DNN) based on Long Short-Term Memory (LSTM) for the task of dialogue act recognition, and we compare
its performance to a standard Maximum Entropy model.
Our first objective is to leverage the modelling capacity of such a DNN in order to achieve dialogue act recognition with only the
raw observed word forms, i.e., without any additional expert-designed feature.
This model is described in Section~\ref{sec:lstm}.
The second objective is to further validate this model both on a standard English DA corpus, as well as on
two other languages, without changing anything in the model, in order to assess the genericity and robustness of the approach.
These experiments are summarized in Section~\ref{sec:experiments}.
Finally, our third objective is to study the impact of word embeddings,
which have been shown to provide extremely valuable information
in numerous Natural Language Processing (NLP) tasks, but which have never been used so far~\footnote{To the best of our knowledge at the time of submission} for dialogue act recognition.
This study is summarized in Section~\ref{sec:embed}.

\section{Related work}
\label{sec:rel_work}


Although dialogue act recognition has been extensively studied in English and German, relatively few works have been published for
Czech and French.
This explains why most of the following related works concern English.
Different sets of dialogue acts are defined in the literature,
depending on the target application and available corpora.
James Allen and Mark Core have proposed DAMSL (Discourse Annotation and Markup System of Labeling), a scheme developed primarily for annotation of two-agent task-oriented dialogues with DAs~\cite{Allen97}. 
This scheme has further been adapted by Daniel Jurafsky to create ``SWBD-DAMSL''~\cite{Jurafsky97a}.
The authors describe a shallow discourse tag-set of approximately 60 basic DA tags (plus combinations) to annotate 1,155 5-minute conversations,
including 205,000 utterances and 1.4M words, from the Switchboard corpus of English telephone conversations.
The Meeting Recorder DA (MRDA) tag-set~\cite{Dhillon04}
is another popular tag-set, which is based on the SWBD-DAMSL taxonomy.
MRDA contains 11 general DA labels and 39 specific labels.

These large DA tag-sets are often reduced for recognition into a few broad classes, because
some classes occur rarely,
or because some DAs are not useful for the target application.
A typical grouping may be for instance:

\vskip 1.0em
\begin{itemize}
\itemsep -1.3mm
\item statements
\item questions
\item backchannels
\item incomplete utterance
\item agreements
\item appreciations
\item other
\end{itemize}
\hskip 1.0em

Automatic recognition of dialogue acts is usually achieved using
one of, or a combination of, the following types of information:
\begin{enumerate}
\item lexical (syntactic and semantic) information
\item prosodic information
\item dialogue history
\end{enumerate}

Lexical information (i.e. the words sequence in the utterance) is useful
for automatic DA recognition, 
because different DAs are usually composed of different word sequences.
Some cue words and phrases can thus serve as explicit indicators of dialogue structure.
For example, 88\% of the trigrams
``$<$start$>$ do you'' occur in English in {\em yes/no questions}~\cite{Jurafsky97}.
Several methods, typically based on word unigrams, may be used to represent lexical information~\cite{Stolcke00}.

Syntactic information is related to the {\em order} of the words in the utterance.
For instance, in French and Czech, the relative order of the
{\em subject} and {\em verb} occurrences
might be used to discriminate between declarations and questions.
With $n>1$, word n-grams may also be used to capture some local syntactic information.
Kr\'al et al. propose to represent word position in the
utterance in order to take into account global syntactic information~\cite{kral2007lexical}. 
Their approach gives 95.8\% DA recognition accuracy on Czech train ticket reservation corpus with 4 DA classes.
A recent work in the dialogue act recognition field~\cite{Kral14} also successfully uses a set of syntactic features derived from a deep parse tree.
The reported accuracy is 97.7\% on the same corpus.

A few related works include semantic features for recognizing dialogue acts.
One of these works combines syntactic parsing of sentences with named entity classes to 
achieve DA recognition from audio input~\cite{liang2011semantic}.
The proposed approach achieves 84.3\% DA detection accuracy on the Tainan-city tour-guiding  dialogue  corpus.
Other researchers employ syntactic and semantic relations acquired by information extraction methods with Bayesian network classifier~\cite{kluwer2010using}.
The obtained dialogue act classification accuracies are 73\% on the 804 sentences of the ``CST'' corpus,
and 68.5\% on the 435 sentences of the ``NPC'' corpus; both corpora are labeled with 7 dialogue acts and deal with the task of furnishing a living-room with the help of a sales agent (a Wizard-of-Oz).

Prosodic information~\cite{Shriberg98}, particularly the melody
of the utterance, is often used to provide additional
clues to label sentences with DAs.
Finally, another important information is the ``dialogue history''. It encodes the sequence of previous dialogue acts~\cite{Stolcke00}
and gives 71\% of dialogue act accuracy on the Switchboard DA corpus when combined with lexical and prosodic features.

Dialogue act recognition is usually based on supervised machine learning, such as Bayesian Networks~\cite{Keizer02},
dynamic Bayesian networks~\cite{yahya2010dynamic}, 
BayesNet~\cite{petukhova2011incremental},
Neural Networks~\cite{Levin03}, but also Boosting~\cite{Tur06}, 
Maximum Entropy Models~\cite{Ang05}, Conditional Random Fields~\cite{Quarteroni11}, Triangular-chain CRF~\cite{Jeong08} and probabilistic rules~\cite{lison2015hybrid}.


Despite the growing importance of deep learning architectures in image processing, speech recognition and several
other natural language processing tasks, deep neural models have only rarely been applied so far
to dialogue act recognition.
In one of these works~\cite{Zhou2015408}, a multi-modal deep neural network is used to extract features and compute abstract representations of the input.
Then, a CRF model takes this input to recognize a~sequence of dialogue acts.
This model achieves 77\% of recognition accuracy on the Chinese CASIA-CASSIL corpus.
In the only other work we are aware of that applies deep learning to DA recognition~\cite{deep13},
the authors combine a deep convolutional model with a vanilla recurrent network across sentences.
They report an accuracy of 73.9\% with this model on the Switchboard-DAMSL corpus. 
We propose in this work an alternative approach to model within-sentence sequential dependencies that is based on the LSTM recurrent cell.
An advantage of this model is that the total number of parameters of the model does not depend on the sentence length.
Furthermore, none of both related works use pretrained word embeddings, and we thus explicitely study in this work
the influence of pretrained word embeddings in this deep architecture.

\section{Dialogue act recognition architecture}
We describe next our proposed system architecture, models and dialogue act recognition procedure.

\subsection{Dialogue act recognition}
The dialogue act recognition task is actually composed of two sub-tasks:
one for dialogue act segmentation and another for dialogue act classification.
For practical application of dialogue act recognition systems, it is essential to
perform both tasks, either in cascade~\cite{Ang05} or jointly by
augmenting the DA labels with BIO-like (Begin-Inside-Out) prefixes~\cite{petukhova2011incremental}, or with N-Gram transducers~\cite{martinez2015unsegmented}.
However, it is also a common practice to assume that segmentation is given and to only classify the given segments.
In this work, we focus on the impact of word embeddings for dialogue act classification, and we thus also assume that the segmentation is known,
which allows us to compare our results with both seminal works~\cite{Stolcke00} and other recent neural models~\cite{deep13}. 

Let us assume that the evaluation corpus is composed of $N$ sentences $(s_i)_{1\leq i\leq N}$, where each sentence $s_i$ is composed of $M_i$ words
$s_i=(w_{i,j})_{1\leq j\leq M_i}$.
The dialogue act recognition task objective is to predict the most likely dialogue act $y_i \in \mathcal{DA}$ 
for every sentence $s_i$.
$\mathcal{DA}$ is the set of all possible dialogue acts for a given corpus. 

The objective is thus naturally cast as a classification task, with varying input sizes because the length of sentences varies.
This varying input size is handled in two different ways, depending on the chosen classifier:
\begin{itemize}
\item Our first classifier is a Maximum Entropy (ME) classifier~\cite{berger1996maximum}, which takes as input a fixed number of
features, represented as a vector $X$, and outputs the probability distribution $P(Y=y|X)$ where $y\in \mathcal{DA}$.
This classification approach is popular in the natural language processing community, because it often gives high recognition scores.
\item Our second classifier, described in Section~\ref{sec:lstm}, is based on a recurrent neural network, and can thus theoretically
take as input a varying number of time-dependent feature vectors $X=(X_t)_{1\leq t\leq M_i}$ for each sentence $i$.
It also outputs normalized scores that can be interpreted as a probability distribution $P(Y=y|X)$ over all possible labels.
\end{itemize}

For the first classifier, the input features are simply the word forms in the current sentence, represented as a one-hot encoding
vector and pooled together as a bag-of-words by taking the maximum value for each dimension of the
one-hot vectors over the whole sentence. Hence, the size of $X$ is $|V|$, where $V$ is the vocabulary size and
$X$ is typically sparse, for instance:
$$X=[0,0,1,0,1,0,0,...]^T$$
where every 1 in the vector represents the occurrence of a given word from the vocabulary in the current sentence.
When further including word embeddings, the feature vector $X$ is concatenated with another feature vector that is the average
of all word embeddings vectors in the current sentence.

For the second classifier, the input observations for each sentence are simply the sequence of one-hot encoding word vectors,
which are immediately transformed in the first layer of the model into a sequence of word embeddings.
In addition, a bag-of-words vector is computed over the words of the previous sentence.

In the training stage, every sentence-level gold label integer
$y_i \in \{1,\dots,|\mathcal{DA}|\}$ is transformed into a $|\mathcal{DA}|$-dimensional output vector that contains 0 everywhere
except at dimension $y_i$.
Then, for the first classifier, the cross-entropy is minimized with the Limited-memory Broyden~Fletcher~Goldfarb~Shanno (L-BFGS) algorithm, while for the second
classifier, it is minimized with adaptive stochastic gradient descent with adaptive moment estimation (``adam'').

\subsection{Deep model}
\label{sec:lstm}

\begin{figure}[h]
\begin{center}
\includegraphics[width=14cm]{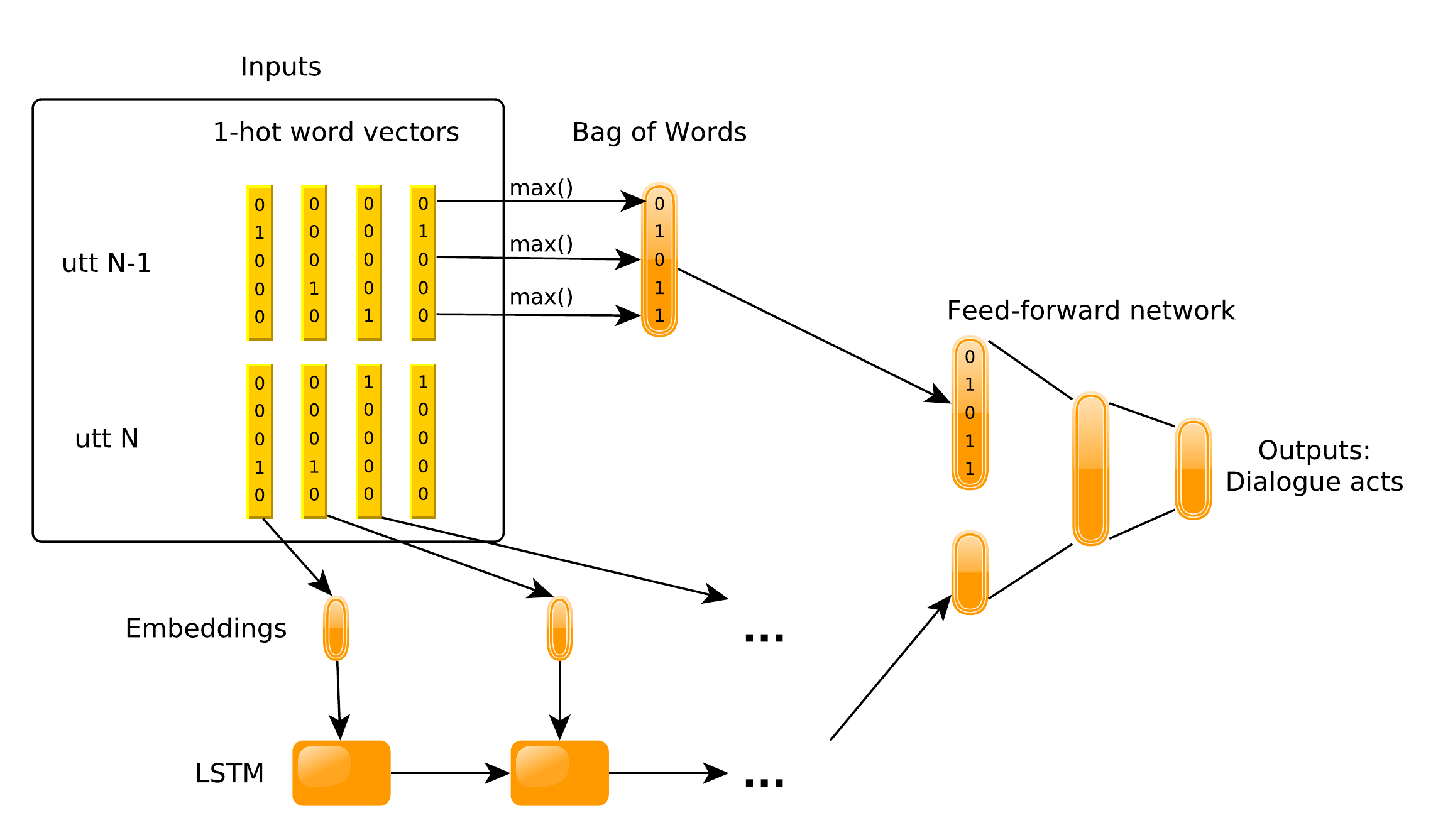}
\end{center}
\caption{Proposed neural network model for dialogue act recognition.}\label{fig:lstm}
\end{figure}

The proposed deep network is shown in Figure~\ref{fig:lstm}.
In the top left of this figure, the input data consists of fixed-length sequences of word indices, one
sequence per sentence.
Every word is represented as a one-hot-encoding word vector $X_t$, i.e., a binary vector with $0$ everywhere
except at a unique word index dimension where it is $1$.

These input word vectors are passed to a first embedding layer, sometimes also called a table look-up layer,
which transforms this one-hot word representation $X_t \in \{0,1\}^{|V|}$ into a smaller real vector $X'_t \in \rm I\!R^{d_e}$ where $d_e << |V|$ is the embedding size.
In the following experiment, before training, this vector is either initialized randomly~\footnote{with uniform distribution between $[-0.05;0.05]$}, or with 
a pretrained word2vec embedding.

These word embeddings are then fed into a recurrent LSTM neural network,
which outputs a single real vector $h \in \rm I\!R^{d_h}$ at the end of each sentence, where $d_h$ is the size of the hidden LSTM state. It is then 
concatenated with another bag-of-words binary vector $Z \in \{0,1\}^{|V|}$, which encodes the observed words of
the previous sentence.
Including the words from the previous sentence compensates for the lack of 
dialogue history model and dynamic programming at the sentence level in our model.

This concatenated vector $\left[ _Z^h \right]$ is finally passed to a two-layer perceptron:
$$u=\tanh \left(W_1 \left[ _Z^h \right] + b_1 \right)$$
$$o=f\left(W_2 u + b_2\right)$$
where $W_1$, $W_2$, $b_1$ and $b_2$ are the parameters and $f(\cdot)$ is the softmax function, which normalizes the output scores so that they may be interpreted as a~probability:
$$f_j(z)=\frac{e^{z_j}}{\sum_{k=1}^{|\mathcal{DA}|} e^{z_k}}$$

\section{Corpora}
\label{sec:corpora}

\subsection{Switchboard dialogue act corpus}
\label{sec:swbd}

The Switchboard-DAMSL corpus~\cite{Jurafsky97a}
is one of the reference English corpora\footnote{http://compprag.christopherpotts.net/swda.html}
to evaluate and compare dialogue
act recognition systems~\cite{Stolcke00,nihms,hk07,margolis10,webb,CSL09}.
Pitifully, many works in the literature publish dialogue act recognition accuracy on this corpus with various experimental
settings, which makes comparison difficult:
\begin{itemize}
\item First, the manual annotation of the corpus has been realized with 60 basic tags plus their combinations. Most works that
report results for dialogue act recognition merge these tags into a smaller tag-set, but which is not always the same: the most frequently
used tag-sets include either 44, 42 or 41 labels, but it is also sometimes reduced to less than 10 dialogue acts.
We have chosen to use the 42 labels that are defined by the corpus authors, in their coder manual\footnote{\url{https://web.stanford.edu/~jurafsky/ws97/manual.august1.html}}.
To obtain these 42 labels, we have chosen to remove (and not merge) all utterances tagged with the '+' segment class in order to obtain the same number of test utterances and number of tokens as in the original paper.
\item Second, the training/test split of the full corpus is often different: some works evaluate the performance with cross-validation over the full corpus, while others randomly extract a list of test files from the full list, and may repeat such a random sampling several times to study variability. We have rather chosen to use the training/test partition defined by the authors of the corpus and available on their website\footnote{\url{http://web.stanford.edu/~jurafsky/ws97}}.
\end{itemize}
We have finally applied the Stanford CoreNLP tokenizer on these utterances, because some of the test sentences were not tokenized in the original corpus.

We have thus made the best possible efforts to reproduce the original experimental setup used in the
seminal paper on this corpus~\cite{Stolcke00} in order to reliable compare our dialogue act recognition scores with related works.
We thus obtain 198,000 training and 4,182 test utterances.
Note that with this experimental setup, the human level of agreement, and so the maximum
recognition accuracy that can be achieved~\cite{Stolcke00} is 84\%.

\subsection{Czech Railways dialogue corpus}
The Czech Railways dialogue corpus
contains human-human conversations in Czech for a train ticket reservation task.
It was recorded at the University of West Bohemia by 28 (14 female and 14 male) speakers being mainly members of the Department of Computer Science and Engineering.
As detailed in Table~\ref{tab:C1}, the corpus contains manual transcriptions of 4,066 spoken utterances, for a total of 26,660 words.
Every utterance has been manually annotated with one out of 9 possible dialogue acts:
statement~(S), order~(O), yes/no question~(Q[y/n]), open question~(Q), yes-answer~(yes), no-answer~(no),
thanks-like sentence~(thank), opening sentence~(hello) and closing sentence~(end).
Additional statistics about the corpus are given in Table~\ref{tab:corpora}.
Table~\ref{tab:C2} shows examples of utterances from this corpus.

\begin{table}[htb]
\begin{center}
\small{
    \begin{tabular}{p{3.5cm} c}
 DA type & Count \\
    \hline\hline order & 125\\
 closing sentence & 204\\
 yes/no question & 282\\ 
 yes-answer  & 308\\
 thanks-like sentence & 406\\
 opening sentence & 434\\
 no-answer & 541\\
statement  & 566\\
open question  & 1,200\\
    \hline
    total & 4,066\\
    \hline
    \end{tabular}
}
\caption{Distribution of dialogue acts in the Czech Railways corpus.}
\label{tab:C1}
\end{center}
\end{table}

\begin{table}[htb]
\begin{center}
\small{
\begin{tabular}{p{1.75cm}p{5.2cm}p{5.2cm}}
DA & Example & English translation \\
\hline\hline
S & Cht\v{e}l bych jet do P\'isku. & I would like to go to P\'isek. \\
O & Najdi dal\v{s}\'i vlak do Plzn\v{e}! & Look at for the next train to Plze\v{n}! \\
Q[y/n] & \v{R}ekl byste n\'am dal\v{s}\'i spojen\'i? & Do you say next connection? \\
Q & Jak se dostanu do \v{S}umperka? & How can I go to \v{S}umperk? \\
\hline
\end{tabular}
}
\caption{Examples for the Czech Railways dialogue corpus.}
\label{tab:C2}
\end{center}
\end{table}

\subsection{French Emospeech corpus}

The French Emospeech corpus is a corpus of dialogues in the context of a serious game between human players and
Non-Player Characters (NPC), which are controlled by an automatic dialogue manager in the real game.
However, when the Emospeech corpus was recorded, all NPCs were 
controlled by Wizard-of-Oz humans who endorsed the role
of the NPCs and communicated with the players via a text-based chat window integrated in the game GUI.
The dialogues collected cover multi-task human-machine interactions for the serious game Mission 
Plastechnology\footnote{\url{http://www.missions-plastechnologie.com}}.
This game is designed to promote careers in the plastic industry. It is a multi-player quest where the players (3
teenagers) seek to build a video game joystick in order to free their uncle trapped in the game. To build this joystick,
the players must explore the plastic factory and achieve 17 mandatory goals (find the plans, get the appropriate mould,
retrieve some plastic from the storing shed, etc)~\cite{emospeech}.

The collected data consists of 1,200 dialogues, 9,417 utterances and 127,000 words.
The annotation scheme designed for the game combines core communicative acts with domain specific information.
The latter defines the goals being pursued/discussed/achieved etc., while the communicative act can be viewed as specifying how the
 current system state is updated by the speaker’s utterance.
In the following experiment, the original set of dialogue acts has been reduced down to the following 9 dialogue acts:
greeting (greet), closing (quit), ask for help in the game (help), other question (ask), gives information (inform), yes answer (yes),
no answer (no), acknowledgement (ack) and miscellaneous (other). The distribution of these dialogue acts is shown in Table~\ref{tab:CES}.

\begin{table}[htb]
\begin{center}
\small{
  \begin{tabular}{p{1.75cm} c}
    DA type & Count \\
\hline \hline
     other & 236\\
     help & 250\\
     no &  307\\
     ack & 489\\
     yes & 704\\
     greet & 1,412\\
     quit & 1,495\\
     ask & 2,084\\
     inform & 2,440\\
    \hline
    total & 9,417\\
    \hline
    \end{tabular}
}
\caption{Distribution of dialogue acts in the French Emospeech corpus.}
\label{tab:CES}
\end{center}
\end{table}

\subsection{Corpora comparison}
Table~\ref{tab:corpora} compares all three corpora.
Details about the French corpus, especially with regard to the Wizard-Of-Oz procedure used to record the corpus as well as
the inter-annotator agreement can be found in a published paper~\cite{emospeech}.
In particular, the inter-annotator agreement has been computed on the French corpus on a detailed set of 27 labels, while we only
use a reduced set of 9 tags in the following experiments. Hence, the French inter-annotator agreement cannot be compared to
our system accuracy, because it obviously is much greater than 80~\% on 9 labels, although we do not know its exact value on these conditions.
We nevertheless report this 27-labels inter-annotator agreement because it gives a good idea about the quality of the corpus and annotation scheme.
For the Czech corpus, we did not find the original inter-annotator agreement so we estimated it by annotating ourselves 250 randomly
extracted sentences.

Table~\ref{tab:corpora} shows that the corpora significantly differ in terms of statistical properties (number of dialogue acts, size...), tasks (reservation, game playing...)
and languages. 
By evaluating our system on these three corpora, we expect to partly assess its robustness to a variety of experimental conditions.

With regard to statistical properties, the main observations are summarized below:
\begin{itemize}
\item The size of the English corpus is significantly greater than both other corpora, about 20 times than the French corpus;
\item The number of DA labels of both smaller corpora is about 5 times lower than in the Switchboard DAMSL corpus;
\item The Czech corpus is 2 times smaller than the French one, both with regard to corpus size and vocabulary size.
\end{itemize}

\begin{table}[htb]
\begin{center}
\small{
  \begin{tabular}{p{3.5cm} ccc}
Corpus & Switchboard & Czech railways & French corpus \\
\hline
\hline
Language & English & Czech & French \\
Task & Conversations & Reservation & Game playing \\
Nature & Human-human & Human-human & Bot(WoZ)-human \\
Nb of DA instances & 202,182 & 4,066 & 9,417 \\
Vocabulary size & 42,000 & 1,203 & 2,294 \\
Number of DA labels & 42 & 9 & 9 \\
Inter-annotator &&&\\
agreement & 0.84 & 0.984 & 0.80 \\
($=\frac{nb~common~labels}{nb~labels}$) &&&\\
\end{tabular}
}
\caption{Main properties of the dialogue corpora used for evaluation.}
\label{tab:corpora}
\end{center}
\end{table}

\section{Experiments}
\label{sec:experiments}

\subsection{Tools and configuration}
\label{sec:Tools}

\subsubsection{Resources availability}

In the following experiments, pretrained word embeddings are either computed or downloaded from the internet:
\begin{itemize}
\item For English, we use the 300-dimensions English word2vec embeddings trained on Google News\footnote{\url{https://code.google.com/archive/p/word2vec}}
\item For French, we use the 200-dimensions French word2vec embeddings trained on FrWac\footnote{\url{http://37.187.110.113/dl/frWac_non_lem_no_postag_no_phrase_200_cbow_cut100.bin}}
\item For Czech, we use the 300-dimensions Czech word2vec embeddings trained on the January 2015 snapshot of Wikipedia.cz
\end{itemize}

The Brainy toolkit~\cite{Konkol:2014} is used to run the Maximum Entropy classifier.

The deep network model is implemented with the Keras deep learning toolkit~\cite{keras}.
The code of the proposed model is available on GitHub\footnote{\url{https://gist.github.com/cerisara/f71e511d594b8c736d65}}.

\subsubsection{Systems configuration}

All experiments in Czech and in French are realized using a 10-fold cross-validation procedure.
Experiments in English on the Swichboard DAMSL corpus are done on a predefined train/test split that matches
the experimental conditions previously published on this corpus, and thus allow for a direct comparison with them,
as explained in Section~\ref{sec:swbd}.

In all our experiments, we only use raw lexical features, i.e. the observed tokenized word forms.
In particular, we do not use part-of-speech tags, nor any other feature, apart from word embeddings.

The number of different labels $|\mathcal{DA}|$ depends on the language and corpus; it is $|\mathcal{DA}|=42$ for English on the
Switchboard-DAMSL corpus, but only $|\mathcal{DA}|=9$ for the French and Czech corpora.

Our vocabulary $V$ contains the 999 most frequent words in the training corpus.
Every rare word that is not in $V$ is replaced by the special word UNK.
So $|V|=1000$. The input feature vector of the Maximum Entropy classifier, which includes both a bag-of-words vector and
the average of all word embeddings in the current sentence is of size $dim(X)=1,300$ for English and Czech, and of size $dim(X)=1,200$ for French.
For our DNN model, the same vocabulary of 1000 words is used and its inputs are shown in Figure~\ref{fig:lstm}.
The hidden state of our LSTM contains 50 neurons and we use 50\%-dropout at the output of the LSTM to regularize training.
The concatenated hidden vector from the LSTM plus the previous bag-of-words vector are passed into a three-layers Multi-Layer Perceptron (MLP)
with 200 hidden neurons and an hyperbolic tangent activation function.
The final output layer has one neuron per dialogue act and is followed by a softmax activation that outputs a value that can be
interpreted as a probability distribution amongst all possible dialogue acts.
This model is trained with the Adam algorithm, which is one of the best performing gradient descent algorithm in many tasks~\cite{adam}\footnote{The hyper-parameters used for Adam are the same as suggested in the original paper, i.e., $\alpha=0.001$, $\beta_1=0.9$, $\beta_2=0.999$ and $\epsilon=10^{-8}$.}.
It minimizes categorical cross-entropy with 20 epochs over the training corpus.
The same model is used in all three languages. The only modification is that French has 200- instead of 300-dimensional embeddings, in order
to comply with the downloaded pretrained embeddings.
All the other model hyper-parameters are language-independent.
Note that gradient descent updates the layers weights and bias from the softmax output down to the embedding weights included,
both when these weights are initialized with pretrained embeddings and randomly.
This is a difference with the Maximum Entropy model, where the pre-trained word vectors are not fine-tuned on the dialogue
act corpus.

Every sentence is truncated or padded with the special word {\em PADDING} so that its length is equal to 15 words.
Having fixed-length sentences is not theoretically required, but it is a common heuristic used in a number of major deep learning libraries (Keras, Theano, Tensorflow, ...) that 
rely on static computation graphs, as opposed to deep learning libraries that dynamically build their computation graph (Torch, DyNet, Tensorflow-Fold, ...), both
paradigms having their respective advantages and drawbacks. For instance, static graphs make variable-length inputs more difficult to handle, but they are easier to
parallelize.
Another reason why this heuristic is often used is that long sentences may be problematic when training an LSTM, because of the
well-known exploding/vanishing gradient issue, but also because of the issue of {\it catastrophic forgetting} in neural networks.
While gradient-clipping and linear pass-through gated connections are suitable solutions to somehow mitigate the former, the latter remains
problematic in recurrent networks~\cite{coop2013mitigation}.
Conversely, truncating and padding may have undesired consequences.
So, when padding a short sentence, the added input elements are commonly not taken into account
during the computation of the loss, thanks to another heuristic called {\it masking}, which applies a mask over the padded elements
that makes their contribution null when computing the loss, and the corresponding weights are thus not trained during back-propagation.
The only potential downside is then the loss of valuable information when truncating sentences.
However, we have shown in a previous work~\cite{kral2007lexical} that, for dialogue act recognition, the most important information is contained at the beginning of
the sentence, and to a lesser extent at the end.
So in our experiments, we systematically copy the last five words of the sentence into positions 10 to 15 of the input sequence, in order to loose as little information as possible.
We show experimentally in Section~\ref{sec:hyper} that with this strategy, the impact of truncating is negligible.

Another potential limitation of this architecture may come from the lack of sentence-level model to capture
typical sequences of dialogue acts. This information is however partly modeled by the bag-of-words input vector from
the previous sentence.
We have realized some additional experiment by rescoring the lattice of probabilities obtained after running our model
onto the sequence of sentences with a smoothed bi-gram of dialogue acts and computing the globally optimal dialogue acts sequence with Viterbi,
but the resulting accuracy gain was not statistically significant.
So this confirms that considering the bag-of-words from the previous sentence may be sufficient for the proposed model in this context.

\subsection{Model hyper-parameters evaluation}
\label{sec:hyper}
In this section, we present a series of experiments performed in order to study the influence of hyper-parameters in our model. 
We use 5-fold cross-validation on the training data from the Switchboard corpus for this purpose.
In each experiment all hyper-parameters are fixed except a single hyper-parameter that is studied.

\begin{figure}[h!]
\begin{center}
\includegraphics[width=12cm]{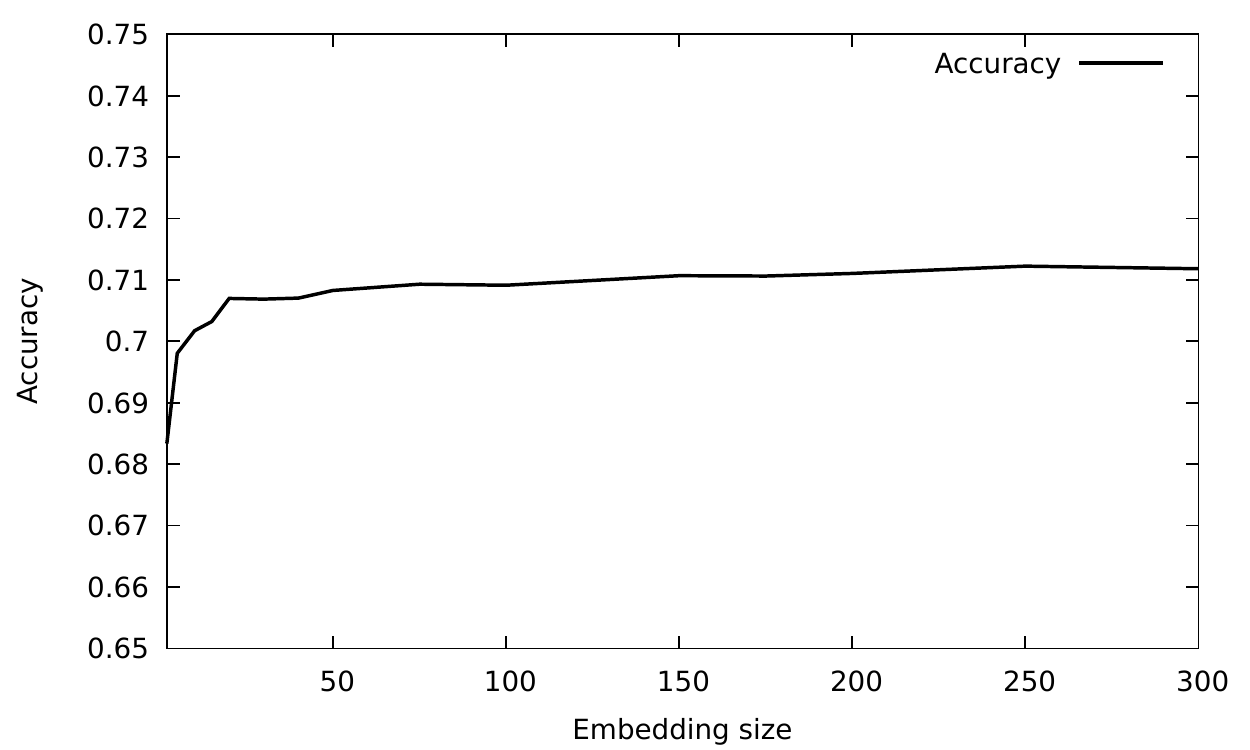}
\caption{Influence of word embeddings size on the DNN accuracy.}
\label{fig:emb}
\end{center}
\end{figure}

The first experiment investigates the impact of word embedding dimension.
All word vectors are randomly initialized in this experiment.
Figure~\ref{fig:emb} shows that reasonable accuracies may be obtained even with 
low-dimensional embeddings.
However, in the following experiments, we nevertheless use 200 and 300-dimensional embeddings because of
the pretrained word vectors used to initialize the embeddings.

\begin{figure}[h!]
\begin{center}
\includegraphics[width=12cm]{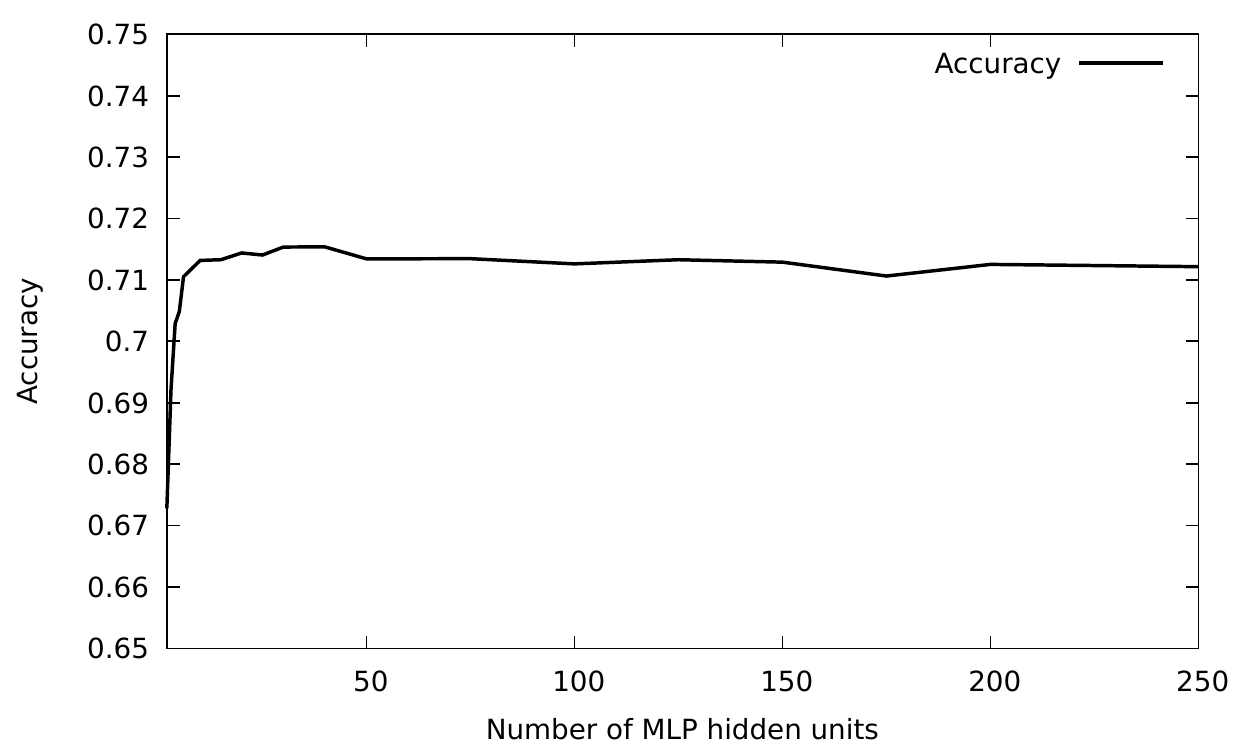}
\caption{Influence of the MLP hidden layer size on the DNN accuracy.}
\label{fig:mlp}
\end{center}
\end{figure}

Figure~\ref{fig:mlp} shows the system accuracy when varying the number of hidden neurons in the MLP.
The highest accuracy is achieved with approximately 40 neurons.
Then the curve becomes very flat and little progress is observed.

\begin{figure}[h!]
\begin{center}
\includegraphics[width=12cm]{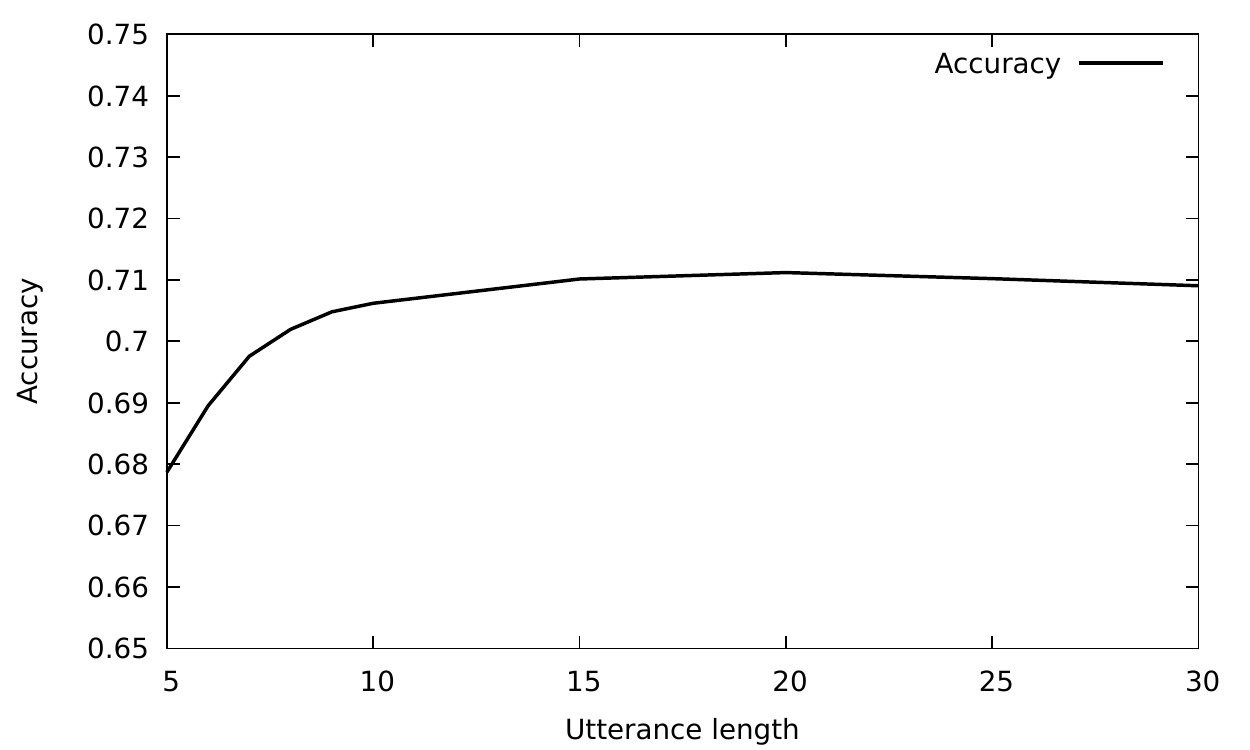}
\caption{Influence of the maximum utterance length on the DNN accuracy.}
\label{fig:utterance}
\end{center}
\end{figure}

Figure~\ref{fig:utterance} depicts the impact of truncating sentences length.
The curve shows that the best performance is reached between 15 and 20 words.
Further increase of the utterance length does not bring any improvement.

\begin{figure}[h!]
\begin{center}
\includegraphics[width=12cm]{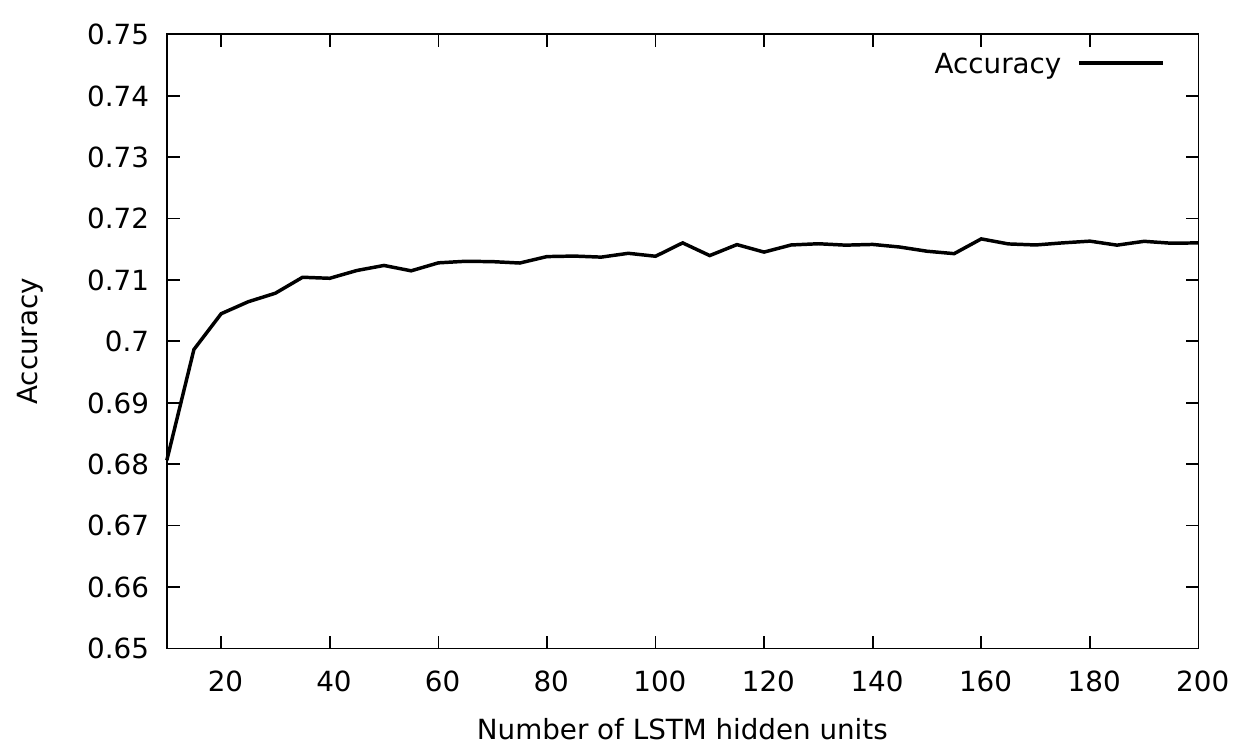}
\caption{Influence of the size of the LSTM hidden state on the DNN accuracy.}
\label{fig:hidden}
\end{center}
\end{figure}

Figure~\ref{fig:hidden} depicts the influence of the size of the LSTM hidden state on the system performance.
We can observe that a reasonable accuracy is achieved at about 50 neurons.
The model tends to achieve slightly better results when increasing this number but the gain is not statistically significant.

\begin{figure}[h]
\begin{center}
\includegraphics[width=12cm]{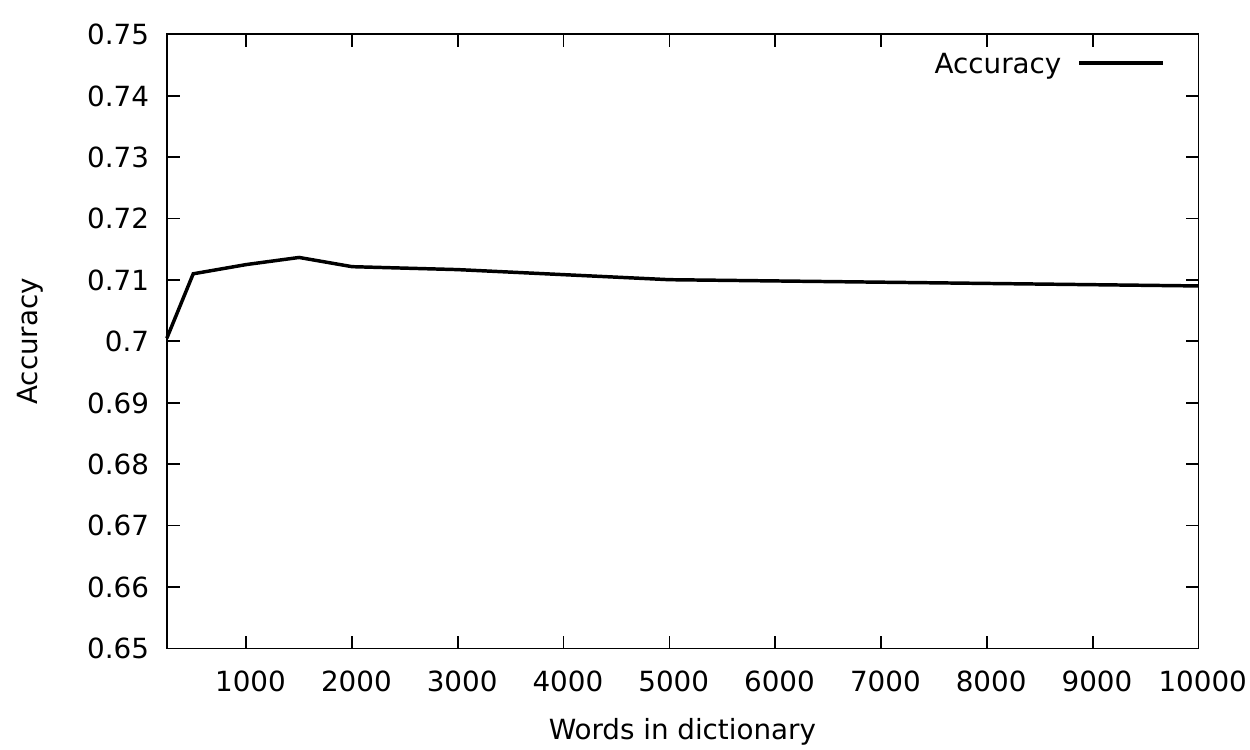}
\caption{Influence of the vocabulary size on the DNN accuracy.}
\label{fig:words}
\end{center}
\end{figure}
 
Figure~\ref{fig:words} shows the evolution of the model accuracy with the size of the vocabulary.
The best accuracy is obtained between 1000 and 2000 words.
Then, the accuracy slightly decreases with larger vocabularies.

\subsection{Dialogue act recognition performance}

We evaluate in this Section the performance of our dialogue act recognition model on three languages: English, Czech and French.
In all experiments,
the confidence interval is computed by assuming a Gaussian distribution of the errors, which leads to the
classical Wald test for a proportion estimation:
$$\hat p \pm z_{1-\frac \alpha 2}\sqrt{\frac 1 n \hat p (1-\hat p)}$$
where $\hat p$ is the accuracy estimate, $n$ the number of examples and $z_{1-\frac \alpha 2}=1.96$ for a 95\%-confidence interval~\cite{confint}.

\subsubsection{Experiments in English}

\begin{figure}[h]
\begin{center}
\includegraphics[width=12cm]{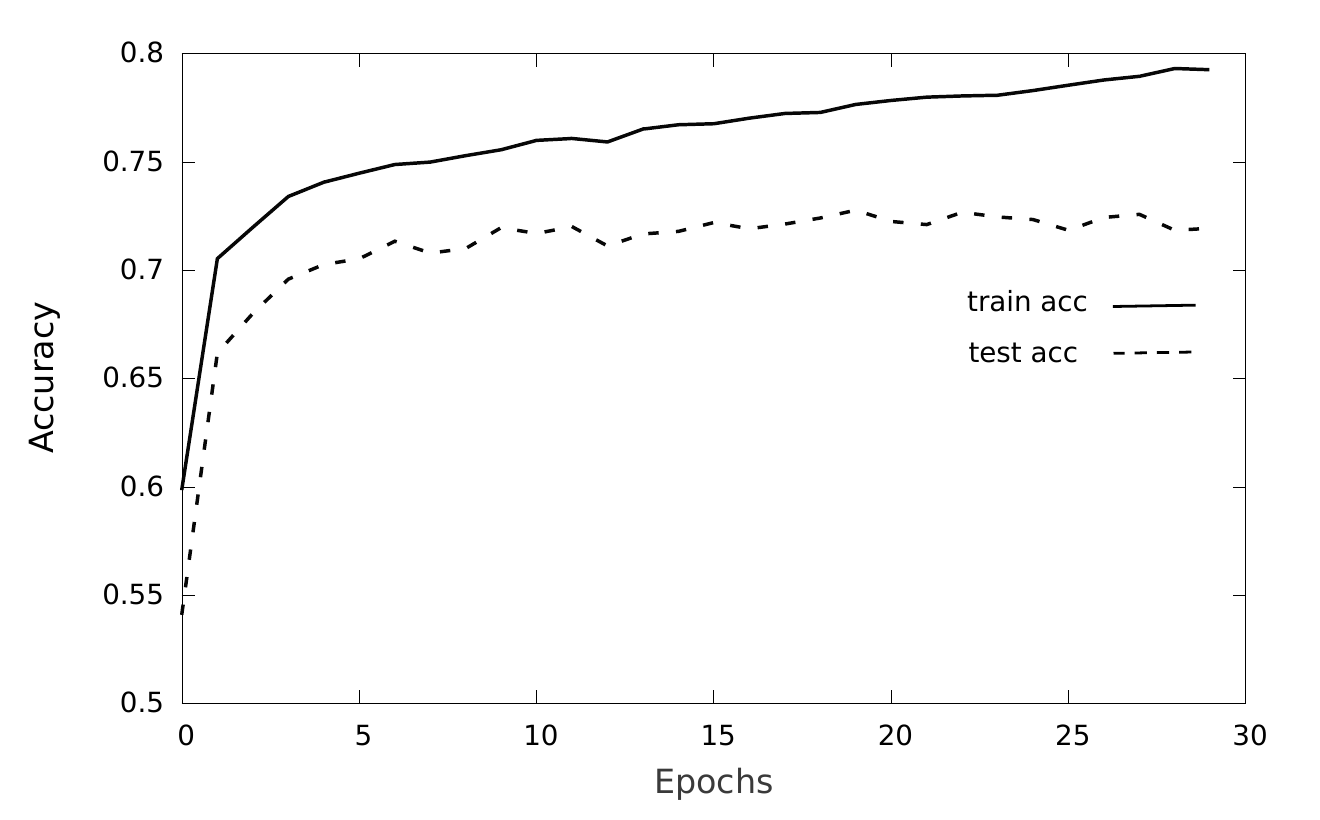}
\caption{Convergence of the training and test accuracy as a function of the number of epochs during training.}
\label{fig:convergence}
\end{center}
\end{figure}

Figure~\ref{fig:convergence} shows the convergence of the accuracy during training of our deep neural network, and also
compares the curves of the accuracy on the training set vs. the accuracy on the test set, which suggests that the model is not overfitting too much.
We can see on this figure that the network converges after about 10 epochs, and its accuracy is then quite stable on the test corpus, with some variations that are due to the relatively small test corpus size: 
the 95\% confidence interval on the Switchboard DAMSL corpus is $\pm 1.35\%$.
The accuracy on the training corpus keeps on slowly increasing with the corpus size, which suggests that
regularization in our model could be slightly enforced.

Table~\ref{tab:resen} reports the accuracy obtained with various configurations of our proposed systems,
as well as some state-of-the-art results.
\begin{table}[h]
\begin{center}
\begin{tabular}{lc}
System & Accuracy \\
\hline \hline
Human level of performance~\cite{Stolcke00} & 84\% \\ 
LM-HMM trigram~\cite{Stolcke00} & 71.0\% \\
Deep network~\cite{deep13} with speaker IDs + dialogue model & 73.9\% \\
\hline
Maximum Entropy (One-hot Bag of Word) & 67.4\% \\
Maximum Entropy (Google News word2vec embeddings) & 61.5\% \\
Maximum Entropy (Oracle embeddings) & 66.5\% \\
Maximum Entropy (BoW + Google News word2vec embeddings) & 69.1\% \\
Maximum Entropy (BoW + oracle embeddings) & 69.3\% \\
Deep neural network (random initialization of the embeddings) & {\bf 72.8}\% \\
Deep neural network (Google News word2vec embeddings) & 72.5\% \\
\hline
\end{tabular}
\caption{Performance of the proposed systems on the English Switchboard DAMSL corpus. Oracle embeddings are the embeddings obtained after training of the deep neural network.}\label{tab:resen}
\end{center}
\end{table}


The best accuracy reported in the literature at the time of submission of this work is 73.9\% \cite{deep13}, which is also obtained with a deep neural network.
However, this accuracy is obtained with an additional input information, the speaker ID.
Our deep model still obtains a level of performance 
that is comparable to the state-of-the-art~\cite{deep13}, given that the difference between both results is not statistically
significant. Conversely, the improvement obtained when replacing the Maximum Entropy model by the deep neural model is
statistically significant. We can thus conclude that deep neural networks give better results than Maximum Entropy models,
and that word embeddings may help to somehow compensate for the weakness of Maximum Entropy models.

We also performed an experiment with 11-fold cross-validation as proposed in a related work~\cite{martinez2015unsegmented}, 
which gives an accuracy of 72.0\% that can be compared to the HMM baseline reported in the paper that gives 70.5\%. 
Although this 11-fold partition of the Switchboard corpus is not yet considered as a standard one, it presents the
advantage of enabling much more precise comparisons between systems thanks to a smaller confidence interval, in our case $\pm 0.2\%$.

\subsubsection{Experiments in Czech}

Table~\ref{tab:XP1} shows the accuracy of dialogue act recognition experiments on the Czech corpus.

\begin{table}[!htb]
\begin{center}
\begin{tabular}{lc}
System & Accuracy \\
\hline
\hline
Maximum Entropy (One-hot Bag of Word) & 96.9\% \\
Maximum Entropy (BoW + word2vec embeddings) & 97.7\% \\
Deep neural network (random initialization of the embeddings) & 98.3\% \\
Deep neural network (word2vec embeddings) & 97.9\% \\
\hline
\end{tabular}
\caption{Dialogue acts recognition accuracy for different approaches/classifiers on the Czech corpus, 95\% confidence interval: $\pm 0.43\%$.}
\label{tab:XP1}
\end{center}
\end{table}
We observe that there is no statistically significant difference between the Maximum Entropy and DNN classifiers
on the Czech corpus.
As suggested by our own evaluation of the inter-annotator agreement on the Czech corpus, the maximum entropy model actually already achieves
the best possible performances, and there is no room left for the more complex model to outperform it.
Furthermore, given our qualitative analysis of some sentences in the Czech corpus, we explain these good performances of the maximum entropy model by the fact that these sentences
are rather simple and follow relatively standard syntactic patterns.

\subsubsection{Experiments in French}

Table~\ref{tab:XP2} shows the accuracy of dialogue act recognition experiments on the French corpus.

\begin{table}[!htb]
\begin{center}
\begin{tabular}{lc}
System & Accuracy \\
\hline
\hline
Maximum Entropy (One-hot Bag of Word) & 86.5\% \\
Maximum Entropy (BoW + word2vec embeddings) & 87.3\% \\
Deep neural network (random initialization of the embeddings) & 92.8\% \\
Deep neural network (word2vec embeddings) & 92.7\% \\
\hline
\end{tabular}
\caption{Dialogue acts recognition accuracy for different approaches/classifiers on the French corpus, 95\% confidence interval: $\pm 0.68\%$.}
\label{tab:XP2}
\end{center}
\end{table}
We observe that the DNN classifiers clearly outperform the Maximum Entropy classifier on the French corpus.
Furthermore, pretrained word2vec embeddings seem to be useful with the Maximum Entropy classifier, but are not
helping the deep neural network.
This observation is confirmed also in Czech and English, and word embeddings are thus further studied
in Section~\ref{sec:embed}.

\subsection{Impact of the training corpus size}
\label{ft:oracle}

We have shown so far that initializing the embeddings with word2vec vectors 
does not help dialogue act recognition on the tested corpora.
We propose in this section three different hypothesis to explain this fact, and study their respective
validity.

\begin{figure}[h]
\begin{center}
\includegraphics[width=12cm]{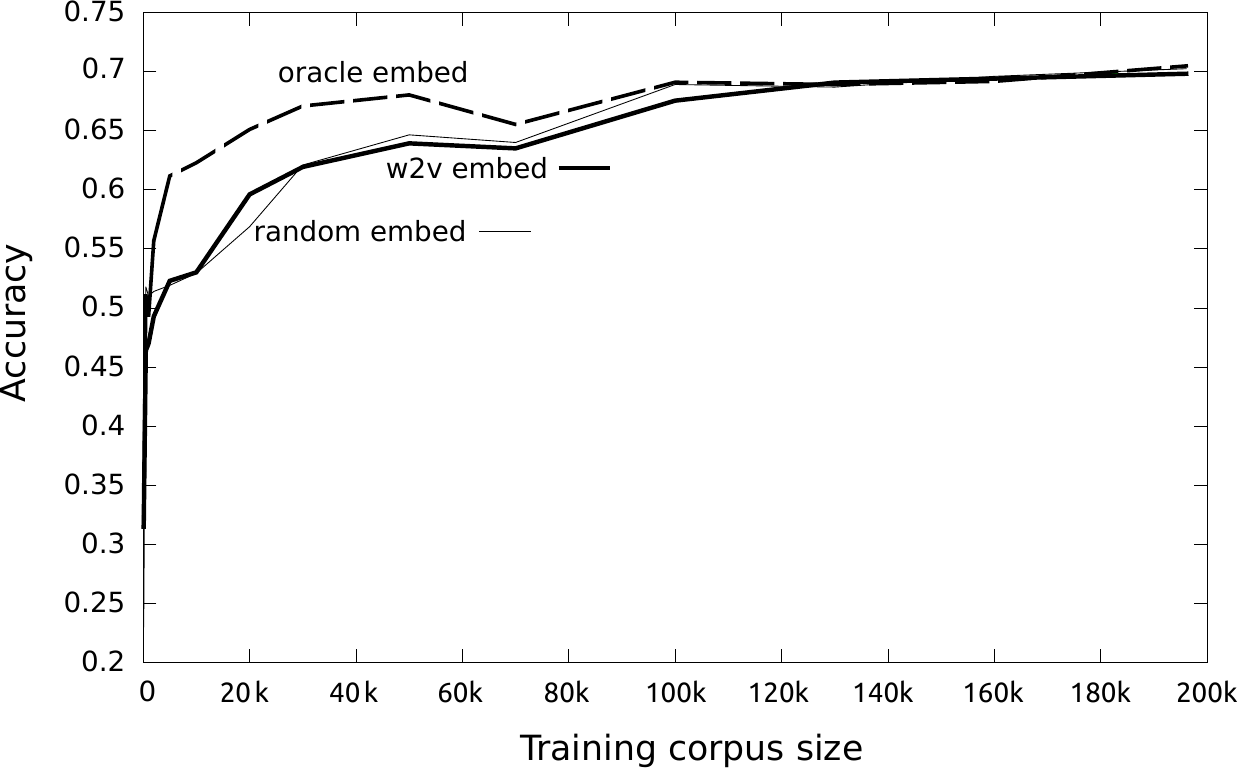}
\caption{Curve of the accuracy as a function of the size of the training corpus. The x-axis is the size of the training corpus
in number of sentences, and the y-axis is the accuracy.}
\label{fig:curve}
\end{center}
\end{figure}

The first hypothesis is that there is enough training data for the embeddings to converge near a global optimum
from every initial position in the parameter space, random or not.
We study this hypothesis by training our model on an increasing corpus size.
Figure~\ref{fig:curve} shows the resulting accuracy curve for three models:
\begin{itemize}
\item {\em Random embed} is for randomly initialized embeddings with a uniform distribution
\item {\em w2v embed} is for embeddings initialized with the Google News word2vec
\item {\em Oracle embed} is for embeddings initialized with the final embeddings obtained after training the model for 20 epochs
over the whole corpus.
\end{itemize}
The accuracy on Figure~\ref{fig:curve} does not converge to exactly the same accuracy as reported in Table~\ref{tab:resen},
because it is obtained with only five epochs of training, in order to reduce experimental time.

We can observe that initializing the embeddings with the Google News word2vec does not improve the dialogue act recognition
accuracy, even for a~small size of the training corpus.
This suggests that our hypothesis that a~large amount of training data may
compensate for a poor random initialization of the embeddings does not hold.

Another possible reason is that the embeddings weights are actually irrelevant, because there is enough parameters
in the rest of the model to capture all the information required to reach this level of accuracy.
In order to study this hypothesis, we initialize the embeddings randomly and train the rest of the model, keeping the
embeddings fixed. The resulting accuracy is only 69.7\%, which is well below the accuracy obtained when training 
the embeddings (72.8\%). So, this invalidates our second hypothesis.

The third possible reason is that,
despite the large success of such embeddings for many NLP tasks,
the standard Google News word2vec does not perform well for the dialogue act recognition task.
Indeed, we note that these types of embeddings were never used in any of the previous works about dialogue act recognition. 
Furthermore, we observe in Figure~\ref{fig:curve} that there exist
some embeddings, called {\em oracle} in this figure\footnote{{\em Oracle} embeddings are ``normal'' embeddings obtained after training on the whole corpus. No additional information is used to compute them.},
which perform better for small corpus sizes.
In particular, this suggests that the standard word2vec embeddings trained on Google News may not be very well adapted to the dialogue act recognition task.
We confirm this assumption by further examining these oracle embeddings in Section~\ref{sec:embed}.

\subsection{Study of the resulting embeddings}
\label{sec:embed}

The next three examples illustrate some typical and frequent mistakes made by our model with and without initialization of the word embeddings with word2vec:

\begin{itemize}
\item ``{\it and I don't think we ever will}'' is correctly recognized as statement-opinion with random embeddings, but as statement-non-opinion with word2vec embeddings
\item ``{\it yeah}'' is correctly recognized as a yes-answer with random embeddings, but as a backchannel with word2vec embeddings
\item ``{\it what should they get in return , I wonder .}'' is correctly recognized as a rhetorical question with random embeddings, but as an open question with word2vec embeddings
\end{itemize}

Please note however that it is difficult to know whether these mistakes are a consequence of different initializations of the embeddings
or whether they are simply due to the noise and natural variability of learnt parameters.
So we study next the embeddings that are obtained after training the deep model for 20 epochs and then extracting the embedding layer.

Table~\ref{tab:yesembed} lists the ten closest words from the word {\em yes} within the embeddings space after
training our deep model for 20 epochs. It also lists the ten closest words according to the Google News word2vec embeddings.
\begin{table}[h]
\begin{center}
\begin{tabular}{ll}
Embed. trained with our model & Google News word2vec \\
\hline \hline
yes & yes \\
yeah & Yes \\
Yep & yeah \\
right & Oh \\
absolutely & oh \\
Yes & hey \\
Absolutely & Yeah \\
agree & suppose \\
true & Nope \\
exactly & guess \\
\hline
\end{tabular}
\caption{Ten closest words from 'yes' according to the cosine distance.}\label{tab:yesembed}
\end{center}
\end{table}

The trained embeddings make a clear distinction between {\em yes} and {\em no}, whereas {\em yes} and {\em nope}
are close in the Google News embeddings. Intuitively, this is understandable, but this lack of fine discrimination between
affirmative and negative answers clearly impact dialogue act recognition accuracy.
Table~\ref{tab:negs} further shows the cosine similarities between {\em yes} and the negative words {\em no}, {\em not} and {\em never}
with both the Google News word2vec embeddings and our trained embeddings.
\begin{table}[h]
\begin{center}
\begin{tabular}{cccc}
 & no & not & never\\
\hline
Google News Word2Vec & 0.392 &  0.391 &  0.355 \\
Our embeddings & 0.292 &  -0.004 &  -0.011 \\
\end{tabular}
\caption{Cosine similarities between {\em yes} and negative words}\label{tab:negs}
\end{center}
\end{table}
On average, these negative words are much closer from {\em yes} with the Google News word2vec (average similarity is $0.380$
and rank is $71$) than with our trained embeddings (average similarity is $0.092$ and rank is $318$).

Similarly, Table~\ref{tab:whatembed} lists the ten closest words from the word {\em what}.
\begin{table}[h]
\begin{center}
\begin{tabular}{ll}
Embed. trained with our model & Google News word2vec \\
\hline \hline
what & what \\
how & exactly \\
What & how \\
who & What \\
Where & something \\
where & why \\
why & everything \\
When & really \\
minutes & anything \\
Huh & know \\
\hline
\end{tabular}
\caption{Ten closest words from 'what' according to the cosine distance.}\label{tab:whatembed}
\end{center}
\end{table}

We can also see that trained embeddings most strongly associate the potentially interrogative pronoun {\em what} with other
similar interrogative pronouns that lead to open answers.
This is also true for Google News word2vec embeddings, but to a lesser extent, as {\em what} is also strongly associated
with content words that are closely semantically related, such as {\em something},{\em anything} and {\em everything}, which
is less relevant with regard to dialogue acts.

The previous experiments as well as these two examples suggest that the word2vec embeddings may not be well adapted
to the dialogue act recognition task.

\section{Conclusions}

We propose in this work an LSTM-based deep neural network for dialogue act recognition.
We show that this model performs as good as the state-of-the-art, even though it only uses the raw word forms as inputs,
without any additional information, in particular neither part-of-speech tags nor information about the speaker.
We have further applied exactly the same model
with the same hyper-parameters on three different languages: English, French and Czech.
The proposed model performs well on all three languages, suggesting that its performance generalizes nicely to
various types of corpora and is not dependent on a specific tuning of the hyper-parameters to experimental conditions.
This confirms the interesting modelling potential of deep recurrent networks for NLP in general, and supports the conclusions
of recent works in the domain~\cite{deep13}, which demonstrate the good performance of end-to-end training of
deep neural networks for dialogue act recognition.

A more surprising conclusion of our work concerns the actual impact of pretrained word embeddings, which have been shown
to be of great importance in several NLP tasks in the literature.
We show in this work that standard pretrained embeddings do not help for the dialogue act recognition task in any of the
three tested languages.
We thus further study the embeddings that result from training the proposed model in an end-to-end manner, and show that they
seem to differ from vanilla word2vec embeddings, which may explain why they do not perform as well as in other tasks.
Of course, a single type of word embeddings has been tested in this work, word2vec, but some additional preliminary experiments
suggest that LDA and COALS-based embeddings do not help either. More experiments with various embeddings should be made to
confirm or infirm this conclusion, but it would be more convincing if they were realized with another deep network implementation
and in more variable experimental conditions.
To the best of our knowledge, this is the first work that exploits pretrained word embeddings for dialogue act recognition, and
one of the rare published work that shows and analyzes some weakness of word2vec embeddings.

We further compare the proposed deep neural network with a standard Maximum Entropy classifier, and show that the DNN
consistenly outperforms the Maximum Entropy classifier both in French and English. This is not the case in Czech, but
it is likely due to the already high level of accuracy reached on this corpus, which leaves little to be gained
by improving the model.
A more interesting conclusion about this comparison between DNN and Maximum Entropy is that 
pretrained word embeddings improve the Maximum Entropy model but not the DNN.
This likely results from the limited modelling capacity of the Maximum Entropy model, which still
benefits from the information brought by pretrained embeddings.
But this information is not precise enough for the DNN, as shown in our qualitative analysis of word2vec.

\section{Acknowledgements}
This work has been partly supported by the project LO1506 of the Czech Ministry of Education, Youth and Sports. 
The authors would like to thank Nvidia for the donation of a Titan X GPU card that has been used to run some experiments in this work.
Some experiments were also carried out using the
Grid'5000 testbed, supported by a scientific interest group hosted by
Inria and including CNRS, RENATER and several Universities as well as
other organizations (see \url{https://www.grid5000.fr}).

\bibliographystyle{splncs03}

\bibliography{others,pavel,dialog_acts,paper}

\begin{thebibliography}{10}
\providecommand{\url}[1]{\texttt{#1}}
\providecommand{\urlprefix}{URL }

\bibitem{Allen97}
Allen, J., Core, M.: Draft of {D}amsl: {D}ialog {A}ct {M}arkup in {S}everal
  {L}ayers. In: \url{http://www.cs.rochester.edu/
  research/cisd/resources/damsl/RevisedManual/RevisedManual. html} (1997)

\bibitem{Ang05}
Ang, J., Liu, Y., Shriberg, E.: Automatic dialog act segmentation and
  classification in multiparty meetings. In: Proceedings of the Acoustics,
  Speech, and Signal Processing Conference (ICASSP). vol.~1, pp. 1061--1064.
  IEEE (Mar 2005)

\bibitem{Austin62}
Austin, J.L.: How to do {T}hings with {W}ords. Clarendon Press, Oxford (1962)

\bibitem{berger1996maximum}
Berger, A.L., Pietra, V.J.D., Pietra, S.A.D.: A maximum entropy approach to
  natural language processing. Computational linguistics  22(1),  39--71 (1996)

\bibitem{Bunt94}
Bunt, H.: Context and {D}ialogue {C}ontrol. Think Quarterly  3,  19--31 (May
  1994)

\bibitem{keras}
Chollet, F.: Keras. \url{https://github.com/fchollet/keras} (2015)

\bibitem{coop2013mitigation}
Coop, R., Arel, I.: Mitigation of catastrophic forgetting in recurrent neural
  networks using a fixed expansion layer. In: Neural Networks (IJCNN), The 2013
  International Joint Conference on. pp. 1--7. IEEE (2013)

\bibitem{Dhillon04}
Dhillon, R., S., B., Carvey, H., E., S.: Meeting {R}ecorder {P}roject: {D}ialog
  {A}ct {L}abeling {G}uide. Tech. Rep. TR-04-002, International Computer
  Science Institute (February 2004)

\bibitem{fivsel2007machine}
Fi{\v{s}}el, M.: Machine learning techniques in dialogue act recognition. Eesti
  Rakenduslingvistika {\"U}hingu aastaraamat  3,  117--134 (2007)

\bibitem{fukada1998probabilistic}
Fukada, T., Koll, D., Waibel, A., Tanigaki, K.: Probabilistic dialogue act
  extraction for concept based multilingual translation systems. In: Fifth
  International Conference on Spoken Language Processing. vol.~6, pp.
  2771--2774 (1998)

\bibitem{Garner96}
Garner, P.N., Browning, S.R., Moore, R.K., Russel, R.J.: A {T}heory of {W}ord
  {F}requencies and its {A}pplication to {D}ialogue {M}ove {R}ecognition. In:
  ICSLP'96. vol.~3, pp. 1880--1883. Philadelphia, USA (1996)

\bibitem{Jeong08}
Jeong, M., Lee, G.G.: Triangular-chain conditional random fields. IEEE trans.
  on Audio, Speech, and Language Processing  16(7),  1287--1302 (Sep 2008)

\bibitem{Jurafsky97a}
Jurafsky, D., Shriberg, E., Biasca, D.: Switchboard {SWBD-DAMSL}
  {S}hallow-{D}iscourse-{F}unction {A}nnotation ({C}oders {M}anual, {D}raft
  13). Tech. Rep. 97-01, University of Colorado, Institute of Cognitive Science
  (1997)

\bibitem{Jurafsky97}
Jurafsky, D., Bates, R., Coccaro, N., Martin, R., Meteer, M., Ries, K.,
  Shriberg, E., Stolcke, A., Taylor, P., Van Ess-Dykema, C.: Automatic
  {D}etection of {D}iscourse {S}tructure for {S}peech {R}ecognition and
  {U}nderstanding. In: IEEE Workshop on Speech Recognition and Understanding.
  pp. 88--95. Santa Barbara (1997)

\bibitem{deep13}
Kalchbrenner, N., Blunsom, P.: Recurrent convolutional neural networks for
  discourse compositionality. CoRR  abs/1306.3584 (2013),
  \url{http://arxiv.org/abs/1306.3584}

\bibitem{Keizer02}
Keizer, S., R., A., Nijholt, A.: Dialogue {A}ct {R}ecognition with {B}ayesian
  {N}etworks for {D}utch {D}ialogues. In: 3rd ACL/SIGdial Workshop on Discourse
  and Dialogue. pp. 88--94. Philadelphia, USA (July 2002)

\bibitem{adam}
Kingma, D., Ba, J.: Adam: A method for stochastic optimization. arXiv preprint
  arXiv:1412.6980  (2014)

\bibitem{kluwer2010using}
Kl{\"u}wer, T., Uszkoreit, H., Xu, F.: Using syntactic and semantic based
  relations for dialogue act recognition. In: Proceedings of the 23rd
  International Conference on Computational Linguistics: Posters. pp. 570--578.
  Association for Computational Linguistics (2010)

\bibitem{Konkol:2014}
Konkol, M.: Brainy: A machine learning library. In: Rutkowski, L., Korytkowski,
  M., Scherer, R., Tadeusiewicz, R., Zadeh, L.A., Zurada, J.M. (eds.)
  Artificial Intelligence and Soft Computing, Lecture Notes in Computer
  Science, vol. 8468. Springer Berlin Heidelberg (2014)

\bibitem{Kral14}
Kr\'al, P., Cerisara, C.: Automatic dialogue act recognition with syntactic
  features. Language Resources and Evaluation  48(3),  419--441 (8 February
  2014)

\bibitem{kral2007lexical}
Kr{\'a}l, P., Cerisara, C., Kleckov{\'a}, J.: Lexical structure for dialogue
  act recognition. Journal of Multimedia  2(3),  1--8 (2007)

\bibitem{hk07}
Lan, K.C., Ho, K.S., Pong Luk~and, R.W., Leong, H.V.: Dialogue act recognition
  using maximum entropy. Journal of the American Society for Information
  Science and Technology  59(6),  859--874 (2008),
  \url{http://dx.doi.org/10.1002/asi.20777}

\bibitem{Levin03}
Levin, L., Langley, C., Lavie, A., Gates, D., Wallace, D., Peterson, K.: Domain
  {S}pecific {S}peech {A}cts for {S}poken {L}anguage {T}ranslation. In: 4th
  SIGdial Workshop on Discourse and Dialogue. pp. 208--217. Sapporo, Japan
  (2003)

\bibitem{liang2011semantic}
Liang, W.B., Wu, C.H., Chen, C.P.: Semantic information and derivation rules
  for robust dialogue act detection in a spoken dialogue system. In:
  Proceedings of the 49th Annual Meeting of the Association for Computational
  Linguistics: Human Language Technologies: short papers-Volume 2. pp.
  603--608. Association for Computational Linguistics (2011)

\bibitem{lison2015hybrid}
Lison, P.: A hybrid approach to dialogue management based on probabilistic
  rules. Computer Speech \& Language  34(1),  232--255 (2015)

\bibitem{margolis10}
Margolis, A., Livescu, K., Ostendorf, M.: Domain adaptation with unlabeled data
  for dialog act tagging. In: Proceedings of the 2010 Workshop on Domain
  Adaptation for Natural Language Processing. pp. 45--52. Association for
  Computational Linguistics (2010)

\bibitem{martinez2015unsegmented}
Mart{\'\i}nez-Hinarejos, C.D., Bened{\'\i}, J.M., Tamarit, V.: Unsegmented
  dialogue act annotation and decoding with n-gram transducers. IEEE/ACM
  Transactions on Audio, Speech, and Language Processing  23(1),  198--211
  (2015)

\bibitem{petukhova2011incremental}
Petukhova, V., Bunt, H.: Incremental dialogue act understanding. In:
  Proceedings of the Ninth International Conference on Computational Semantics.
  pp. 235--244. Association for Computational Linguistics (2011)

\bibitem{Quarteroni11}
Quarteroni, S., Ivanov, A.V., Riccardi, G.: Simultaneous dialog act
  segmentation and classification from human-human spoken conversations. In:
  Acoustics, Speech and Signal Processing (ICASSP), 2011 IEEE International
  Conference on. pp. 5596--5599. IEEE, Prague, Czech Republic (May 2011)

\bibitem{CSL09}
Rangarajan~Sridhar, V.K., Bangalore, S., Narayanan, S.: Combining lexical,
  syntactic and prosodic cues for improved online dialog act tagging. Comput.
  Speech Lang.  23(4),  407--422 (Oct 2009),
  \url{http://dx.doi.org/10.1016/j.csl.2008.12.001}

\bibitem{Ritter2010172}
Ritter, A., Cherry, C., Dolan, B.: Unsupervised modeling of twitter
  conversations. In: NAACL HLT 2010 - Human Language Technologies: The 2010
  Annual Conference of the North American Chapter of the Association for
  Computational Linguistics, Proceedings of the Main Conference. pp. 172--180
  (2010)

\bibitem{emospeech}
Rojas~Barahona, L.M., Lorenzo, A., Gardent, C.: {Building and Exploiting a
  Corpus of Dialog Interactions between French Speaking Virtual and Human
  Agents}. In: Chair), N.C.C., Choukri, K., Declerck, T., Do{\u g}an, M.U.,
  Maegaard, B., Mariani, J., Odijk, J., Piperidis, S. (eds.) {The eighth
  international conference on Language Resources and Evaluation (LREC)}. pp.
  1428--1435. {European Language Resources Association (ELRA)}, Istanbul,
  Turkey (May 2012), \url{https://hal.inria.fr/hal-00726721}

\bibitem{Sanchis02}
Sanchis, E., Castro, M.J.: Dialogue {A}ct {C}onnectionist {D}etection in a
  {S}poken {D}ialogue {S}ystem. In: Second International Conference on Hybrid
  Intelligent Systems (HIS2002). pp. 644--651. IOS Press, Santiago de Chile,
  Chile (1-4 December 2002)

\bibitem{Shriberg98}
Shriberg, E., Bates, R., Stolcke, A., Taylor, P., Jurafsky, D., Ries, K.,
  Coccaro, N., Martin, R., Meteer, M., Van Ess-Dykema, C.: Language and Speech,
  Special Double Issue on Prosody and Conversation, vol.~41, chap. Can
  {P}rosody {A}id the {A}utomatic {C}lassification of {D}ialog {A}cts in
  {C}onversational {S}peech?, pp. 439--487 (1998)

\bibitem{sridhar2009combining}
Sridhar, V.K.R., Bangalore, S., Narayanan, S.: Combining lexical, syntactic and
  prosodic cues for improved online dialog act tagging. Computer Speech \&
  Language  23(4),  407--422 (2009)

\bibitem{nihms}
Sridhar, V.K.R., Narayanan, S., Bangalore, S.: Modeling the intonation of
  discourse segments for improved online dialog act tagging. In: Acoustics,
  Speech and Signal Processing, 2008. ICASSP 2008. IEEE International
  Conference on. pp. 5033--5036. IEEE (2008)

\bibitem{Stolcke00}
Stolcke, A., Ries, K., Coccaro, N., Shriberg, E., Bates, R., Jurafsky, D.,
  Taylor, P., Martin, R., Van Ess-Dykema, C., Meteer, M.: Dialog {A}ct
  {M}odeling for {A}utomatic {T}agging and {R}ecognition of {C}onversational
  {S}peech. In: Computational Linguistics. vol.~26, pp. 339--373 (2000)

\bibitem{Stolcke98}
Stolcke, A., Shriberg, E., Bates, R., Coccaro, N., Jurafsky, D., Martin, R.,
  Meteer, M., Ries, K., Taylor, P., Van Ess-Dykema, C., et~al.: Dialog {A}ct
  {M}odeling for {C}onversational {S}peech. In: AAAI Spring Symp. on Appl.
  Machine Learning to Discourse Processing. pp. 98--105 (1998)

\bibitem{Tur06}
Tur, G., Guz, U., Hakkani-Tur, D.: Model adaptation for dialog act tagging. In:
  Spoken Language Technology Workshop, 2006. IEEE. pp. 94--97. IEEE (2006)

\bibitem{vosoughi2016tweet}
Vosoughi, S., Roy, D.: Tweet acts: A speech act classifier for twitter. arXiv
  preprint arXiv:1605.05156  (2016)

\bibitem{confint}
Wallis, S.A.: Binomial confidence intervals and contingency tests: mathematical
  fundamentals and the evaluation of alternative methods. Journal of
  Quantitative Linguistics  20(3),  178--208 (2013)

\bibitem{webb}
Webb, N., Hepple, M., Wilks, Y.: Error analysis of dialogue act classification.
  In: Text, Speech and Dialogue. pp. 451--458. Springer (2005)

\bibitem{yahya2010dynamic}
Yahya, A.A., Mahmod, R., Ramli, A.R.: Dynamic bayesian networks and variable
  length genetic algorithm for designing cue-based model for dialogue act
  recognition. Computer Speech \& Language  24(2),  190--218 (2010)

\bibitem{zarisheva2015dialog}
Zarisheva, E., Scheffler, T.: Dialog act annotation for twitter conversations.
  In: Proceedings of the 16th Annual Meeting of the Special Interest Group on
  Discourse and Dialogue. pp. 114--123 (2015)

\bibitem{Zhou2015408}
Zhou, Y., Hu, Q., Liu, J., Jia, Y.: Combining heterogeneous deep neural
  networks with conditional random fields for chinese dialogue act recognition.
  Neurocomputing  168,  408 -- 417 (2015),
  \url{http://www.sciencedirect.com/science/article/pii/S0925231215007845}

\bibitem{zimmermanntoward}
Zimmermann, M., Liu, Y., Shriberg, E., Stolcke, A.: Toward joint segmentation
  and classification of dialog acts in multiparty meetings. Machine Learning
  for Multimodal Interaction pp. 187--193 (2006)

\end{thebibliography}

\end{document}